\documentclass[10pt,twocolumn,letterpaper]{article}

\usepackage[pagenumbers]{cvpr} 

\usepackage{xspace}
\usepackage{tabularx}
\newcommand{\modelname}{EvoScene\xspace}

\definecolor{cvprblue}{rgb}{0.21,0.49,0.74}
\usepackage[pagebackref,breaklinks,colorlinks,allcolors=cvprblue]{hyperref}

\def\paperID{14485} 
\def\confName{CVPR}
\def\confYear{2026}

\title{Self-Evolving 3D Scene Generation from a Single Image}

\author{
    Kaizhi Zheng$^{1}$\quad
    Yue Fan$^{1}$ \quad
    Jing Gu$^{1}$ \quad
    Zishuo Xu$^{1}$ \quad
    Xuehai He$^{1}$ \quad
    Xin Eric Wang$^{1,2}$ \\
    $^{1}$University of California, Santa Cruz \quad
    $^{2}$University of California, Santa Barbara
}

\begin{document}


\twocolumn[{
\renewcommand\twocolumn[1][]{#1}
\maketitle
\centering
\vspace{-1em}
\includegraphics[width=0.95\textwidth]{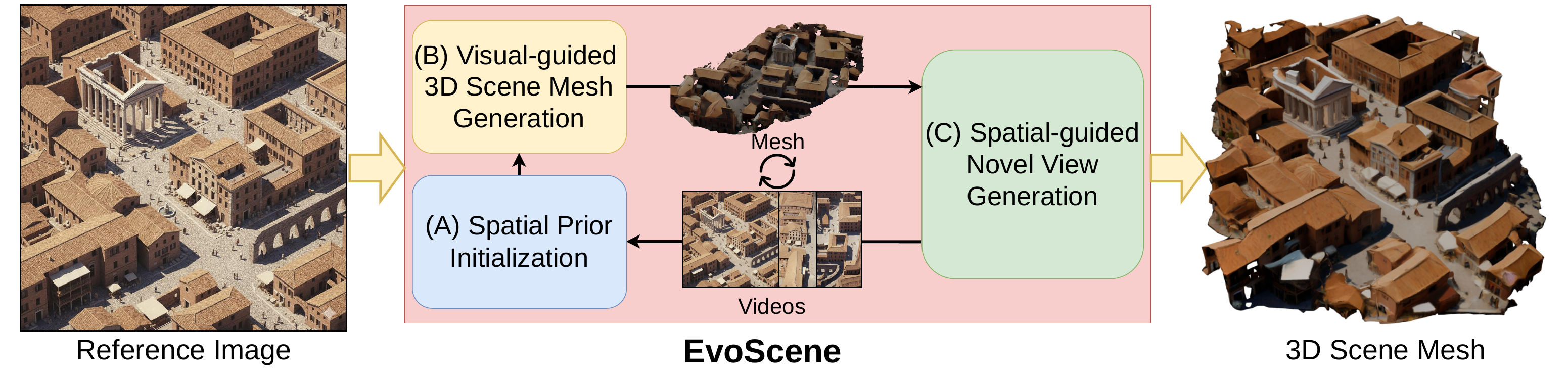}
\hfill\captionof{figure}{\textbf{Self-evolving 3D scene generation from a single image.} Given a single input photograph, \modelname progressively builds a complete 3D scene through a virtuous cycle where geometry and appearance mutually refine each other. Our method synthesizes photorealistic novel views and produces high-quality 3D representations (Gaussians and meshes) with substantially improved angular coverage and completeness. Project page: \url{https://eric-ai-lab.github.io/evoscene.github.io/}}
\vspace{1em}
\label{fig:teaser} 
}]

\begin{abstract}

Generating high-quality, textured 3D scenes from a single image remains a fundamental challenge in vision and graphics. Recent image-to-3D generators recover reasonable geometry from single views, but their object-centric training limits generalization to complex, large-scale scenes with faithful structure and texture.
We present \textbf{\modelname}, a self-evolving, training-free framework that progressively reconstructs complete 3D scenes from single images. The key idea is combining the complementary strengths of existing models: geometric reasoning from 3D generation models and visual knowledge from video generation models. Through three iterative stages—Spatial Prior Initialization, Visual-guided 3D Scene Mesh Generation, and Spatial-guided Novel View Generation—\modelname\ alternates between 2D and 3D domains, gradually improving both structure and appearance.
Experiments on diverse scenes demonstrate that \textbf{\modelname} achieves superior geometric stability, view-consistent textures, and unseen-region completion compared to strong baselines, producing ready-to-use 3D meshes for practical applications.
\end{abstract}

\section{Introduction}
\label{sec:intro}


Games, films, animation, simulation, and many other digital applications rely heavily on high-quality 3D visuals. While recent video generation models can take a single image and render visuals as if a full 3D scene existed, their outputs remain purely image-space: they lack consistent geometry, exhibit viewpoint-dependent drift, and cannot be converted into editable or physically grounded 3D assets. In practice, producing such 3D assets still requires extensive manual modeling and texturing, which demands significant artistic effort and is difficult to scale. This makes an automated approach for generating complete 3D scenes from a single image highly desirable.

Despite its appeal, single-image 3D scene generation remains fundamentally challenging. Limited observations constrain coverage: methods~\cite{ren2025gen3c,huang2025voyager,li2025flashworld} that synthesize novel views along a fixed camera path leave large portions of the scene unseen, leading to holes or hallucinated regions with weak geometric grounding. Ensuring high-resolution, globally consistent textures adds further difficulty. Existing image-to-3D generators~\cite{xiang2024structured,hunyuan3d22025tencent,li2025triposg,li2025sparc3d}, predominantly trained on object-level data, struggle when scaled to complex scenes, producing coarse geometry and misaligned or blurry textures. Extensions~\cite{engstler2025syncity,zheng2025constructing,chen2025trellisworld} that operate on overlapping latent patches expand spatial coverage but still suffer from texture seams and inconsistent fine details. End-to-end scene generation models~\cite{wu2024blockfusion,Meng2024LT3SDLT} attempt a direct mapping from images to 3D scenes, but the scarcity of diverse, high-quality scene datasets restricts them to narrow domains and limits their general applicability.

To address these challenges, we propose a self-evolving, iterative pipeline, \textbf{\modelname}, that progressively improves the 3D scene rather than attempting to recover everything from a single pass, shown in Figure~\ref{fig:teaser}. \modelname operates by cycling through three stages:
In Stage A, Spatial Prior Initialization, \modelname\ builds a spatial prior, a coarse geometric representation of the scene, by estimating depth from the 2D scene images (initially the input image) and fusing it into a point cloud. This step provides an initial understanding of the scene’s layout and major structures.
In Stage B, Visual-guided 3D Scene Mesh Generation, the point-cloud prior is lifted into an explicit 3D mesh using an existing 3D generation model. This mesh captures key surfaces and topology, forming a stable geometric scaffold with editable textures.
In Stage C, Spatial-guided Novel View Generation, the reconstructed mesh guides a video generation model to synthesize new multi-view images that reveal unseen regions and supply additional structure and appearance cues. These new views are then fed back into the next iteration, allowing geometry and texture to improve step by step.

The entire pipeline is training-free, combining complementary knowledge already learned by existing models through its stages. The stages that lift 2D observations into 3D structure (Stages A and B) leverage the geometric reasoning of depth estimation and a 3D generation model to produce a consistent scene-level mesh. The stage that converts this mesh into synthesized multi-view imagery (Stage C) draws on the visual and textural knowledge of a video generation model to achieve detailed, view-consistent appearances beyond mesh-resolution limits. Together, these stages provide wide viewpoint coverage, reliable structure, and high-quality textures in a 3D mesh representation ready for downstream use.

Through extensive experiments across diverse scenes, we demonstrate that \modelname produces complete textured 3D scene meshes with superior geometric quality, spatial layout coherence, and photorealistic appearance compared to state-of-the-art image-to-3D baselines, achieving human preference win rates of 78.5\%-90.5\% across all evaluation criteria. Ablations validate the importance of our core design choices, confirming that both iterative geometry-appearance co-evolution and depth-conditioned video guidance are essential for achieving high-quality scene generation. Both quantitative metrics and qualitative comparisons highlight \modelname's ability to generate geometrically consistent scenes with fine-grained details across varying complexity.

Our main contributions are:
\begin{itemize}
\setlength\itemsep{-0.1em}
\item We propose \textbf{\modelname}, a self-evolving framework for single-image 3D scene generation that progressively expands spatial coverage and refines complete textured meshes through iterative cycles between geometry and appearance across 2D and 3D domains.
\item We introduce a three-stage modular design that synergistically combines point-cloud spatial priors, SLAT-based 3D diffusion for mesh generation, and depth-conditioned video generation, enabling effective integration of geometric knowledge from 3D models with visual priors from video models.
\item Comprehensive experiments with human preference and GPT-4o-based evaluations on diverse scenes demonstrate that \modelname\ achieves superior geometric completeness, layout coherence, and texture fidelity compared to state-of-the-art image-to-3D baselines, with ablations confirming substantial improvements over pose-only world-model variants.
\end{itemize}
\section{Related Work}

\noindent \textbf{Image-to-3D Scene Generation}
While object-level image-to-3D methods~\cite{Poole2022DreamFusionTU,Lin2022Magic3DHT,Tang2023MakeIt3DH3,Liu2023Zero1to3ZO,Tang2023DreamGaussianGG,tang2024lgm,xiang2024structured,hunyuan3d22025tencent,li2025triposg,li2025sparc3d} achieve impressive reconstruction quality, they struggle when scaled to complex scenes due to limited spatial coverage. Scene-level generation requires handling larger spatial extents and maintaining global coherence. Neural scene representations like NeRF~\cite{mildenhall2021nerf} and 3D Gaussians~\cite{kerbl20233d} provide powerful rendering frameworks. Recent approaches leverage 2D diffusion models~\cite{rombach2022high,ho2020denoising,chang2024flux,ramesh2021zero} to outpaint novel views and reconstruct via depth fusion or volumetric representations. Early work focused on indoor scenes~\cite{wiles2020synsin,koh2021pathdreamer,hollein2023text2room,koh2023simple}, with later methods tackling natural scenes~\cite{li2022infinitenature,cai2023diffdreamer,fridman2023scenescape,chung2023luciddreamer,yu2024wonderjourney,zhang20243d,zhang2024text2nerf,yang2024scene123,shriram2024realmdreamer,engstler2024invisible,yu2024wonderworld}. Panoramic~\cite{stan2023ldm3d,li2024scenedreamer360,wu2023panodiffusion,schult2024controlroom3d,liang2024luciddreamer} and multi-view generation~\cite{liu2024panofree,tang2023emergent,gao2024cat3d} improve coverage but typically restrict camera motion and struggle with geometric accuracy due to hallucinations in occluded regions. Alternative approaches directly generate scenes in native 3D through layout prediction~\cite{feng2023layoutgpt,zhou2024gala3d,ccelen2024design,hu2024scenecraft,sun20233d,tang2024diffuscene,lin2024instructscene,zhai2024echoscene,vilesov2023cg3d,maillard2024debara} or latent 3D spaces~\cite{wu2024blockfusion,Meng2024LT3SDLT,Ren2023XCubeL3,Liu2023PyramidDF,Lee2024SemCitySS,Chai2023PersistentNA}, but often require large domain-specific datasets and struggle with generalization. There are some recent works~\cite{engstler2025syncity,zheng2025constructing,yoon2025extendd,chen2025trellisworld,li2025voxhammer} that apply object-level 3D generators to scene-level mesh synthesis without training. Critically, while these methods achieve impressive single-pass results, they lack mechanisms to progressively refine and expand coverage through iterative multi-view observations, leading to incomplete spatial coverage and limited detail in initially unseen areas. Our self-evolving approach addresses this through geometry-appearance co-evolution across multiple iterations.

\noindent \textbf{Video Generation for 3D Scene Reconstruction}
Video diffusion models~\cite{wan2025,HaCohen2024LTXVideo,kong2024hunyuanvideo,liu2024sora,hu2023videocontrolnet,jiang2025vace,peng2024controlnext} capture powerful visual priors for generating photorealistic imagery with temporal coherence. World models extend this to novel view synthesis by conditioning on camera trajectories~\cite{gao2024cat3d,wang2024motionctrl,zhou2025stable,he2024cameractrl}, enabling controllable viewpoint generation. Video generation with 3D priors~\cite{ren2025gen3c,huang2025voyager,li2025flashworld,chen2025mess,kim2025videofrom3d} further incorporates depth or geometry conditioning to improve multi-view consistency. However, these approaches face key limitations: world models with initial depth conditioning~\cite{ren2025gen3c,huang2025voyager,li2025flashworld} are constrained by predefined camera trajectories that limit full scene coverage, and lack mechanisms to accumulate scene knowledge for iterative refinement. Mesh texturing methods~\cite{chen2025mess,kim2025videofrom3d} require pre-existing complete 3D meshes rather than performing reconstruction. Unlike these approaches that use video generation as a final rendering step, we leverage depth-conditioned video diffusion within an iterative reconstruction framework. Our method progressively refines and expands scene geometry through generated multi-view observations, accumulating spatial knowledge across iterations to achieve complete coverage. As our ablations demonstrate, this strategy substantially outperforms pose-only world models, producing geometrically consistent results suitable for 3D mesh reconstruction.

\noindent \textbf{Iterative and Self-Evolving Methods}
Iterative refinement is widely used in 3D generation. Multi-stage pipelines separate geometry and texture generation~\cite{chen2023fantasia3d,metzer2022latent,li2024craftsman3d,yang2024hi3d}, while iterative optimization methods alternate between shape and appearance~\cite{lee2024disr,ryu2025elevating}. Recent work explicitly closes the loop between 2D and 3D domains: Ouroboros3D~\cite{wen2025ouroboros3d} integrates multi-view diffusion and reconstruction in a recursive diffusion process, Cycle3D~\cite{tang2025cycle3d} alternates between 2D generation and 3D reconstruction, GeoDream~\cite{ma2023geodream} disentangles diffusion and cost-volume priors for iterative refinement, and GenFusion~\cite{wu2025genfusion} leverages reconstruction-driven video diffusion. However, these methods rely primarily on multi-view generation models for geometric reasoning, lacking explicit integration of 3D generative priors that encode structural knowledge. Unlike prior methods, our self-evolving framework synergistically combines knowledge from both 3D generation models (for geometric structure) and video generation models (for photorealistic appearance): each iteration uses a 3D diffusion model to complete scene geometry, then leverages this completed mesh to guide video synthesis for texture generation, progressively refining both modalities through their iterative interaction to achieve complete scene generation from a single input image.
\section{Method}
\label{sec:method}

\begin{figure*}[t]
    \centering
    \includegraphics[width=\linewidth]{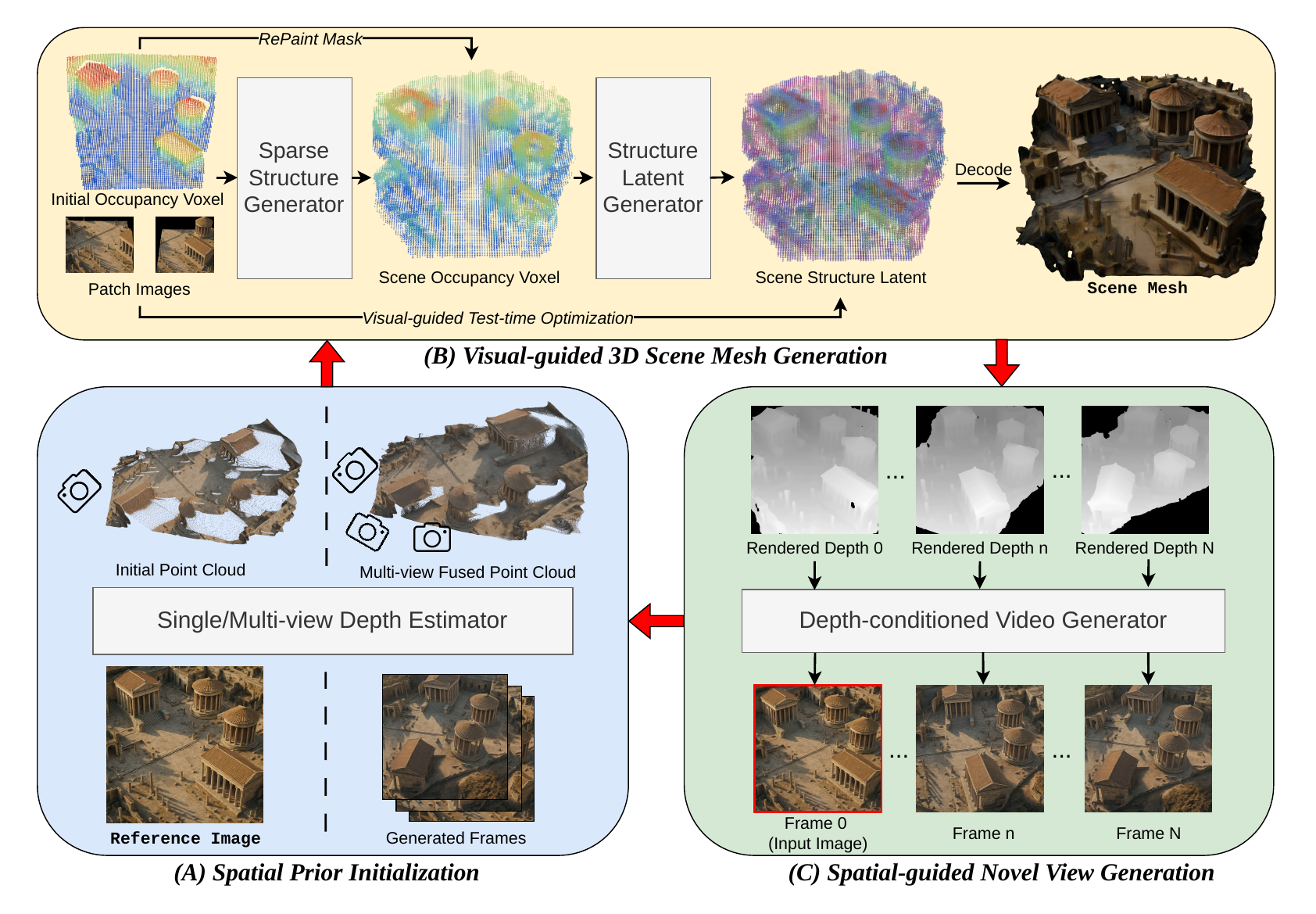}
    \vspace{-2.5em}
    \caption{\textbf{Overview of \modelname.}
    Our self-evolving framework consists of three coupled stages that form a virtuous cycle: (A) \emph{Spatial Prior Initialization} extracts depth from 2D observations and back-projects to 3D point clouds, providing geometric constraints. (B) \emph{Visual-guided 3D Scene Mesh Generation} uses a 3D diffusion model with test-time rendering optimization to complete the mesh guided by point cloud priors. (C) \emph{Spatial-guided Novel View Generation} renders depth from the mesh to guide depth-conditioned video diffusion, synthesizing photorealistic multi-view images. These new views are converted back to point clouds (via depth re-estimation and multi-view filtering) and fed into the next iteration. This cycle progressively refines geometry and appearance to produce a complete 3D scene from a single input image.}
    \label{fig:overview}
    \vspace{-1.5em}
\end{figure*}

\subsection{Overview}
Given a single input image $I_0$, our goal is to generate a complete and view-consistent 3D scene representation. Single-image generation is challenging due to occlusions and depth ambiguity. We leverage two complementary sources of prior knowledge: pretrained \emph{3D generation models} that encode geometric reasoning for plausible structure, and pretrained \emph{video generation models} that capture visual knowledge for realistic textures. Neither alone is sufficient, as geometry-based approaches struggle with high-fidelity textures, while video-based approaches lack explicit 3D consistency.

\modelname\ addresses this through a self-evolving framework with three coupled stages forming a virtuous cycle (Figure~\ref{fig:overview}). \emph{Stage A: Spatial Prior Initialization} extracts geometric priors by estimating depth maps and back-projecting them into point clouds. \emph{Stage B: Visual-guided 3D Scene Mesh Generation} leverages these priors with a pretrained 3D diffusion model to generate a complete mesh filling occluded regions. \emph{Stage C: Spatial-guided Novel View Generation} uses the mesh to guide a video generation model, synthesizing photorealistic multi-view images from novel viewpoints. These newly generated views are converted back into geometric priors through depth re-estimation and fed into the next iteration. Through this cycle, geometry and appearance mutually refine each other, progressively evolving a single image into a complete 3D scene.

\subsection{Self-Evolving 3D Scene Generation Stages}

Let $t$ index the iteration number ($t=0,1,\ldots,T$), $S$ denote voxel resolution, and $P$ denote patch size. At each iteration, we maintain three key data structures that evolve throughout the process:

\noindent(\emph{i}) \textbf{Multi-view observations} $\mathcal{V}_t = \{(I_k, C_k)\}_{k=1}^{N_t}$, where $I_k$ are RGB images and $C_k$ are camera parameters. Initially, $\mathcal{V}_0 = \{(I_0, C_0)\}$ contains only the input image. New views are added in each iteration from Stage C.

\noindent(\emph{ii}) \textbf{Point cloud prior} $\mathcal{P}_t$ with per-point confidence scores, derived from depth estimation and multi-view filtering. This provides sparse, metric-scale geometric constraints extracted from the accumulated observations.

\noindent(\emph{iii}) \textbf{Latent 3D representation} $\mathcal{G}_t$ encoded as SLAT~\cite{xiang2024structured} (Structured LATent), a sparse voxel grid where each occupied voxel stores a latent feature vector. This representation can be decoded into explicit geometries (meshes, 3D Gaussians) for rendering, optimization, and editing.

Each iteration consists of three coupled stages that transform data between these representations:

\paragraph{(A) Spatial Prior Initialization.}
To build a 3D geometric understanding from 2D observations, we extract depth information and lift it into 3D space. In the first iteration ($t=0$), we have only the input image $I_0$. We apply a pretrained monocular depth estimator to obtain a depth map $d_0$. Using the estimated camera intrinsics $K_0$ and extrinsics $E_0$, we back-project the depth map to obtain an initial point cloud:
\begin{equation}
\mathcal{P}_0 = \text{BackProject}(d_0, K_0, E_0).
\end{equation}
Each point is assigned a confidence score based on local depth gradient magnitude, reflecting the reliability of the depth estimation. Points in regions with smooth depth transitions receive higher confidence, while those near depth discontinuities or uncertain areas receive lower confidence.

In subsequent iterations ($t > 0$), we have access to multi-view observations $\mathcal{V}_t = \{(I_k, C_k)\}_{k=1}^{N_t}$ accumulated from previous cycles. For each view, we re-estimate depth maps $\{d_k\}$ and back-project them into point clouds. To ensure geometric consistency and filter out erroneous depth estimates, we apply multi-view geometric voting: for each candidate 3D point, we project it into all available views and check depth agreement. Only points with sufficient cross-view support are retained. The filtered points from all views are merged to form the updated point cloud:
\begin{equation}
\mathcal{P}_t = \mathcal{P}_{t-1} \cup \text{Filter}\Big(\bigcup_{k=1}^{N_t} \text{BackProject}(d_k, K_k, E_k)\Big),
\label{eq:point_cloud_update}
\end{equation}
where $\text{Filter}(\cdot)$ denotes the multi-view voting procedure that removes outliers based on cross-view depth consistency.

This accumulated point cloud $\mathcal{P}_t$ serves as the spatial prior for the next stage, providing metric-scale geometric constraints that anchor the 3D generation process.

\paragraph{(B) Visual-guided 3D Scene Mesh Generation.}
While the point cloud $\mathcal{P}_t$ from Stage A provides sparse geometric constraints, it remains incomplete due to occlusions and limited viewpoint coverage. Large regions of the scene are unobserved, and the point cloud lacks surface topology. To address this, we leverage a pretrained 3D diffusion model to complete the geometry and generate a full 3D mesh representation.

We voxelize the point cloud $\mathcal{P}_t$ at resolution $S^3$ to obtain a sparse occupancy grid $\mathcal{O}_t \in \{0,1,2\}^{S^3}$, where values indicate observed regions (from $\mathcal{P}_t$), carved free-space (ray visibility), and unknown occluded regions to be hallucinated. To handle large scenes, we decompose the grid into overlapping $P^3$ patches $\{\mathcal{B}_j\}$ and extract corresponding image crops $\{I_j\}$ from accumulated observations.

We employ a two-stage flow-matching diffusion generator based on SLAT~\cite{xiang2024structured}:
We first use a sparse structure generator to denoise a binary occupancy latent $Z_t^{\text{SS}}$ with RePaint-style~\cite{lugmayr2022repaint} masking: observed voxels are re-injected from the ground truth encoding while unknown voxels evolve freely, conditioned on image crops $\{I_j\}$:
\begin{equation}
\frac{d}{dt}Z_t^{\text{SS}} = v^{\text{SS}}(Z_t^{\text{SS}}, \{I_j\}, t), \quad \text{s.t. } Z_t^{\text{SS}}|_{\mathcal{O}_t=2} = \mathcal{E}(\mathcal{O}_t),
\end{equation}
where $\mathcal{E}$ encodes the occupancy grid and $v^{\text{SS}}$ is the learned velocity field. This yields a completed occupancy $\hat{\mathcal{O}}_0$ filling occluded regions while preserving observed structure.

Next, we employ a structured latent generator to generate latent features $\{\mathbf{z}_i\}$ for each occupied voxel, given $\hat{\mathcal{O}}_0$ and image crops $\{I_j\}$, by denoising $Z_t^{\text{SLAT}}$. Inspired by Extend3D~\cite{yoon2025extendd}, we apply test-time optimization using multi-view rendering feedback. We periodically decode the current SLAT into 3D Gaussians, render from all accumulated views $\mathcal{V}_t$, and minimize the multi-view reconstruction loss:
\begin{equation}
\begin{split}
\mathcal{L}_{\text{mv}} = \sum_{k=1}^{N_t} \Big[&\lambda_1\|\Pi(\mathcal{G}_t; C_k) - I_k\|_1 + \lambda_2\text{LPIPS}(\Pi(\mathcal{G}_t; C_k), I_k) \\
&+ \lambda_3(1 - \text{SSIM}(\Pi(\mathcal{G}_t; C_k), I_k))\Big],
\end{split}
\label{eq:render}
\end{equation}
where $\Pi(\cdot; C_k)$ denotes the differentiable rendering operator and $\lambda_1, \lambda_2, \lambda_3$ are loss weights (1 by default). The gradient $\nabla_{Z_t^{\text{SLAT}}}\mathcal{L}_{\text{mv}}$ guides denoising toward photometrically consistent geometry.

The output $\mathcal{G}_t$ (encoded as SLAT) can be decoded into 3D Gaussians for differentiable rendering and an explicit mesh (via Marching Cubes) for editing, providing a scaffold for the next stage.
\paragraph{(C) Spatial-guided Novel View Generation.}
While Stage B produces a geometrically complete mesh, its appearance quality is limited by the 3D diffusion model's training data and resolution. To overcome this and expand viewpoint coverage, we leverage the mesh $\mathcal{G}_t$ as a spatial scaffold to guide a pretrained video diffusion model in synthesizing photorealistic multi-view images.

We render $N$ novel viewpoints along an orbital camera trajectory, obtaining depth maps $\{D^{(k)}\}_{k=1}^N$ from the mesh geometry. These depth maps encode the 3D spatial layout, providing structural guidance for video generation. We feed them to a pretrained depth-conditioned video diffusion model with the input image $I_0$ as the first frame and an optional text prompt $p$:
\begin{equation}
\mathcal{V}_{t+1} = \text{VideoDiffusion}\big(I_0, \{D^{(k)}\}_{k=1}^N, p\big).
\label{eq:video_gen}
\end{equation}
The model generates temporally coherent frames $\mathcal{V}_{t+1} = \{I_k^{\text{new}}\}_{k=1}^N$ that are photorealistic and geometrically consistent, hallucinating plausible textures for previously occluded regions.

These synthesized images serve dual purposes: (\emph{i}) they provide appearance supervision for refining mesh texture, and (\emph{ii}) they are fed back into Stage A of the next iteration, where depth re-estimation converts them into geometric priors. This closes the self-evolving loop, enabling geometry and appearance to mutually refine each other across iterations.

\begin{table*}[t]
    \centering
    \caption{\textbf{Quantitative comparisons of \modelname and baselines.} We report human preference win rates and GPT-4o-based weighted win rates (\%) across geometry, layout, and texture quality. \modelname consistently outperforms all baselines by large margins in both evaluations.}
    \resizebox{\textwidth}{!}{
    \begin{tabular}{lcccccc}
        \toprule
        & \multicolumn{3}{c}{Human Preference Win Rate (Percentage)} & \multicolumn{3}{c}{GPT-4o-based Weighted Win Rate (Percentage)} \\
        \cmidrule(lr){2-4} \cmidrule(lr){5-7}
        Models & Geometry Quality & Layout Coherence & Texture Coherence & Geometry Quality & Layout Coherence & Texture Coherence\\
        \midrule
        Trellis~\citep{xiang2024structured} & 19.50\% & 16.00\% & 20.00\% & ~~9.47\% & ~~6.62\% & ~~7.75\%\\
        \modelname (Ours) & \textbf{80.50\%} & \textbf{84.00\%} & \textbf{80.00\%} & \textbf{90.53\%} & \textbf{93.38}\% & \textbf{92.25\%}\\
        \midrule
        Hunyuan3D-2.1~\citep{hunyuan3d22025tencent} & 16.50\% & 11.50\% & 15.50\% & ~~9.72\% & ~~5.13\% & ~~6.42\%\\
        \modelname (Ours) & \textbf{83.50\%} & \textbf{88.50\%} & \textbf{84.50\%} & \textbf{90.28\%} & \textbf{94.87}\% & \textbf{93.58\%}\\
        \midrule
        TripoSG~\citep{li2025triposg} & 15.00\% & ~~9.50\% & 14.00\% & ~9.60\% & ~~6.12\% & ~9.28\% \\
        \modelname (Ours) & \textbf{85.00\%} & \textbf{90.50\%} & \textbf{86.00\%} & \textbf{90.40\%} & \textbf{92.88\%} & \textbf{90.72\%}\\
        \bottomrule
    \end{tabular}
    }
    \label{table:main_result}
     \vspace{-1em}
\end{table*}

\paragraph{Self-evolution schedule.}
The three stages form a \emph{self-evolving cycle} where geometry and appearance co-evolve through iterative refinement. The cycle operates as follows:
\begin{itemize}
\setlength\itemsep{-0.1em}
\item \textbf{Stage A $\to$ B:} Sparse point cloud priors (from depth estimation) provide geometric constraints that anchor the 3D diffusion model, ensuring the completed mesh preserves observed structure.
\item \textbf{Stage B $\to$ C:} The completed mesh provides spatial guidance (via rendered depth maps) that steers video generation toward geometrically consistent, photorealistic appearance.
\item \textbf{Stage C $\to$ A (next iteration):} Synthesized multi-view images are converted back into geometric priors through depth re-estimation and multi-view filtering, providing richer constraints for the next iteration's mesh completion.
\end{itemize}

We iterate this cycle $T$ times ($t=1,\ldots,T$). To maximize scene coverage, each iteration generates videos from complementary camera trajectories (for example, alternating between positive and negative azimuth angles, e.g., $+30^\circ$ in iteration 1, $-30^\circ$ in iteration 2, or varying elevation angles). This strategy progressively expands angular coverage, revealing previously unseen regions and refining both geometry and appearance with each cycle. As iterations progress, the accumulated multi-view observations $\mathcal{V}_t$ grow richer, the point cloud $\mathcal{P}_t$ becomes denser, and the mesh quality improves, ultimately converging to a complete, photorealistic 3D scene.

\paragraph{Mesh extraction.}
After the final iteration, we decode $\mathcal{G}_T$ into both a 3D Gaussian representation and a mesh (via Marching Cubes on the occupancy grid).
We bake multi-view textures onto the mesh using the accumulated observations and export as a GLB file for rendering and editing.

\subsection{Implementation Details}
We employ pretrained models for all components: HunyuanWorld-Mirror~\cite{liu2025worldmirror} for single/multi-view 3D reconstruction, Trellis~\cite{xiang2024structured} for SLAT-based 3D mesh generation, and Wan2.2-Fun-5B-Control~\cite{wan2025} for depth-conditioned video generation.
We set voxel resolution $S=128$ with patch size $P=64$, and render video frames at $1024\times1024$ resolution with 121 frames.
We run $T=3$ self-evolution iterations, generating videos from complementary azimuth angles (e.g., $+30^\circ$ and $-30^\circ$ with uniform view sampling) to progressively expand angular coverage. Every iteration requires 6 minutes with the highest 68G VRAM on one H100 GPU.
All models are used off-the-shelf without fine-tuning. For more method details, please refer to Appendix~\ref{sec:appendix_method}.
\section{Experiments}

\subsection{Experimental Setup}

\paragraph{Benchmark and Metrics.}
We evaluate on 100 diverse scene images generated by GPT-4o~\cite{hurst2024gpt} and GPT-Image-1~\cite{openai_gpt_image_1_docs_2025}. We conduct pairwise comparisons on three criteria: geometry quality, layout coherence, and texture coherence, annotated by two AMT workers (\textbf{human preference win rate}) for each pair and GPT-4o with token probability weighting (\textbf{GPT-4o weighted win rate}). We also report \textbf{CLIP} similarity and \textbf{FID} between rendered views and reference images. More details can be found in Appendix~\ref{sec:appendix_experiment}.

\paragraph{Baselines.}
We compare against three state-of-the-art image-to-3D models: \textit{Trellis}~\cite{xiang2024structured}, \textit{Hunyuan3D-2.1}~\cite{hunyuan3d2025hunyuan3d}, and \textit{TripoSG}~\cite{li2025triposg}. All models are evaluated zero-shot with official checkpoints using complete input images without background removal.

\subsection{Main Results}

\begin{table}[t]
    \centering
    \caption{\textbf{Rendered-view metrics on the reference images.} 
    CLIP similarity and FID are computed between each method's render and the corresponding reference image.}
    \footnotesize 
    \begin{tabularx}{\columnwidth}{l *{2}{>{\centering\arraybackslash}X}}
        \toprule
        Models & CLIP $\uparrow$ & FID $\downarrow$ \\
        \midrule
        \modelname~(Ours) & \textbf{0.8643} & \textbf{188.68} \\
        Trellis~\citep{xiang2024structured} & 0.7454 & 280.67 \\
        Hunyuan3D-2.1~\citep{hunyuan3d22025tencent} & 0.7312 & 322.53 \\
        TripoSG~\citep{li2025triposg} & 0.7152 & 366.34 \\
        \bottomrule
    \end{tabularx}
    \label{table:clip_fid}
     \vspace{-2em}
\end{table}

\begin{figure*}[!t]
     \centering
     \includegraphics[width=\textwidth]{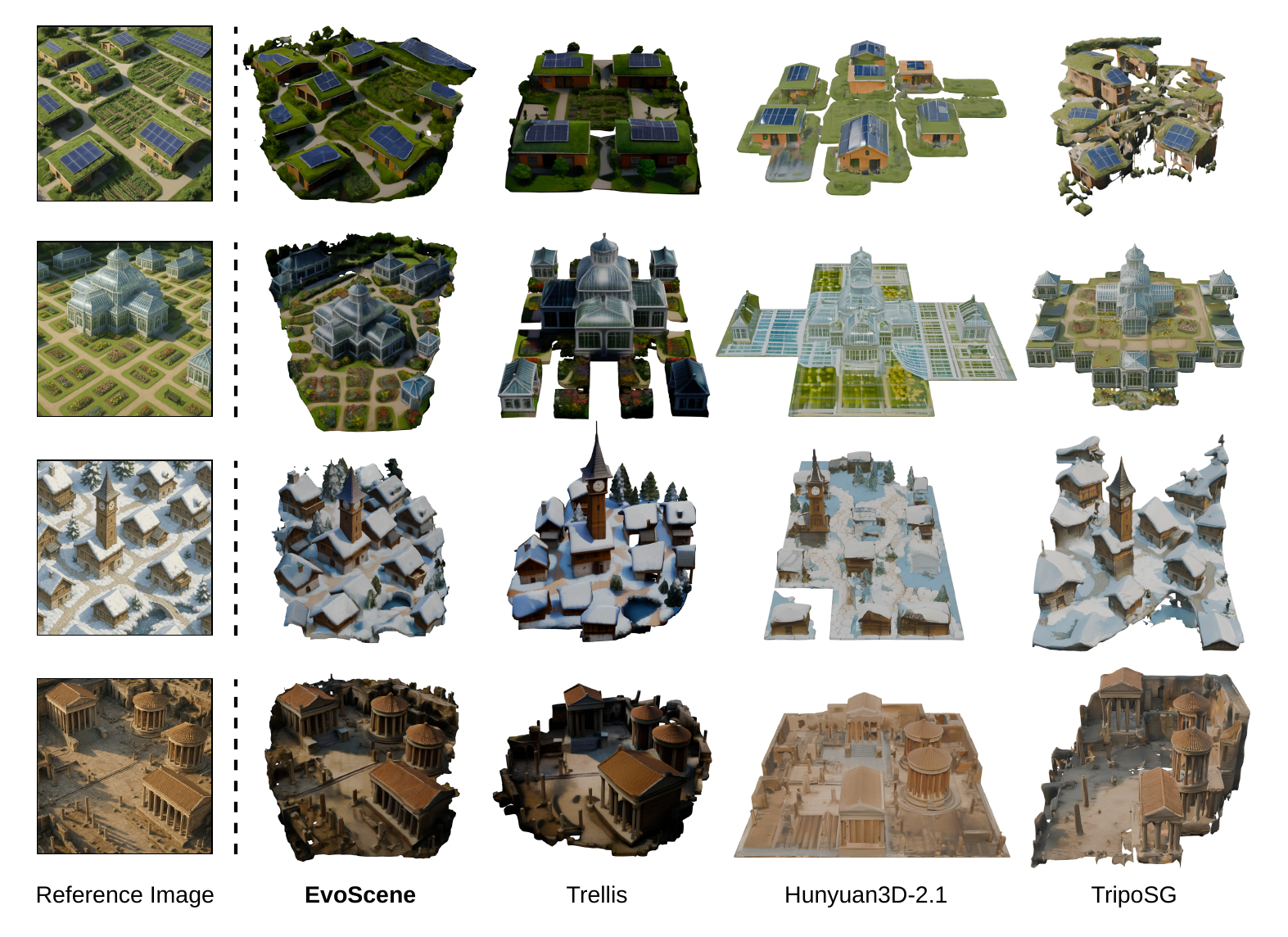}
     \vspace{-2.5em}
     \caption{\textbf{Qualitative comparisons across diverse scene types.} From left to right: reference images, \modelname (ours), Trellis, Hunyuan3D-2.1, and TripoSG. Scenes include residential neighborhoods (rows 1-2), urban architecture (row 3), and historical landmarks (row 4). \modelname produces complete geometry with preserved architectural details and photorealistic textures, while baselines exhibit fragmentation, flattened representations, or severe geometric distortions.}
     \label{fig:qualitative_comparisons}
      \vspace{-0.5em}
\end{figure*}

\begin{table*}[t]
    \centering
    \caption{\textbf{Ablation Study Results.} Human preference and GPT-4o win rates demonstrate the importance of iterative refinement and depth-conditioned video guidance in \modelname.}
    \resizebox{\textwidth}{!}{
    \begin{tabular}{lcccccc}
        \toprule
        & \multicolumn{3}{c}{Human Preference Win Rate (Percentage)} & \multicolumn{3}{c}{GPT-4o-based Weighted Win Rate (Percentage)} \\
        \cmidrule(lr){2-4} \cmidrule(lr){5-7}
        Models & Geometry Quality & Layout Coherence & Texture Coherence & Geometry Quality & Layout Coherence & Texture Coherence\\
        \midrule
        \modelname~(Single Iteration) & 46.00\% & 43.50\%& 47.50\% & 41.78\% & 39.52\% & 47.08\%\\
        \modelname~(Full) & \textbf{54.00\%} & \textbf{56.50\%} & \textbf{52.50\%} & \textbf{58.22\%} & \textbf{60.48\%} & \textbf{52.92\%}\\
        \midrule
        \modelname~(Pose-only Video Gen.) & ~~6.50\% & 10.00\% & ~~6.00\% & ~~1.36\% & ~~0.88\% & ~~2.14\% \\
        \modelname~(Full) & \textbf{93.50\%} & \textbf{90.00\%} & \textbf{94.00\%} & \textbf{98.64\%} & \textbf{99.12\%} & \textbf{97.86\%}\\
        \bottomrule
    \end{tabular}
    }
    \label{table:ablation}
\end{table*}

\paragraph{Quantitative Analysis}
Table~\ref{table:main_result} demonstrates \modelname's substantial advantages across three state-of-the-art baselines. In human preference evaluations, \modelname achieves win rates of 78.5\%-85.0\% for geometry quality, 84.0\%-90.5\% for layout coherence, and 84.5\%-88.0\% for texture coherence. The particularly strong performance in layout coherence indicates our iterative approach effectively preserves spatial relationships and scene structure. GPT-4o-based evaluations show even stronger advantages, with weighted win rates consistently exceeding 90\% across all criteria and baselines, reaching up to 94.9\% for layout coherence against Hunyuan3D-2.1. This suggests that when quality differences are weighted by magnitude, \modelname's superiority becomes more pronounced.

Table~\ref{table:clip_fid} provides automatic metrics measuring semantic fidelity and perceptual quality. \modelname achieves CLIP similarity of 0.8643 (15.9\% improvement over Trellis) and FID of 188.68 (32.8\% improvement), demonstrating superior semantic alignment and visual realism. These results validate our self-evolving approach, where iterative refinement enables geometry and appearance to mutually reinforce each other while progressively expanding scene coverage.

\paragraph{Qualitative Analysis}
Figure~\ref{fig:qualitative_comparisons} provides visual comparisons across diverse scene types. For geometry quality, \modelname produces complete, coherent meshes with well-preserved architectural details including building facades, rooftops, and terrain structures. In contrast, Trellis generates incomplete geometry with missing regions and lacks fine structural details, while TripoSG suffers from severe fragmentation where scene elements appear disconnected. Hunyuan3D-2.1 produces oversimplified geometry missing important features. For layout coherence, \modelname maintains correct spatial relationships with proper depth ordering, while Hunyuan3D-2.1 generates flattened representations where depth relationships are compressed. For texture quality, \modelname synthesizes photorealistic, multi-view consistent appearance through depth-conditioned video generation, exhibiting sharp details and smooth surface transitions. Baselines suffer from various artifacts: Trellis produces blurry textures, TripoSG shows inconsistent coloring, and Hunyuan3D-2.1 lacks realism. \modelname demonstrates consistent quality across varying scene complexity. More experiment results can be found in Appendix~\ref{sec:appendix_experiment}.

\subsection{Ablation Study}

\begin{figure}[t]
    \centering
    \includegraphics[width=\columnwidth]{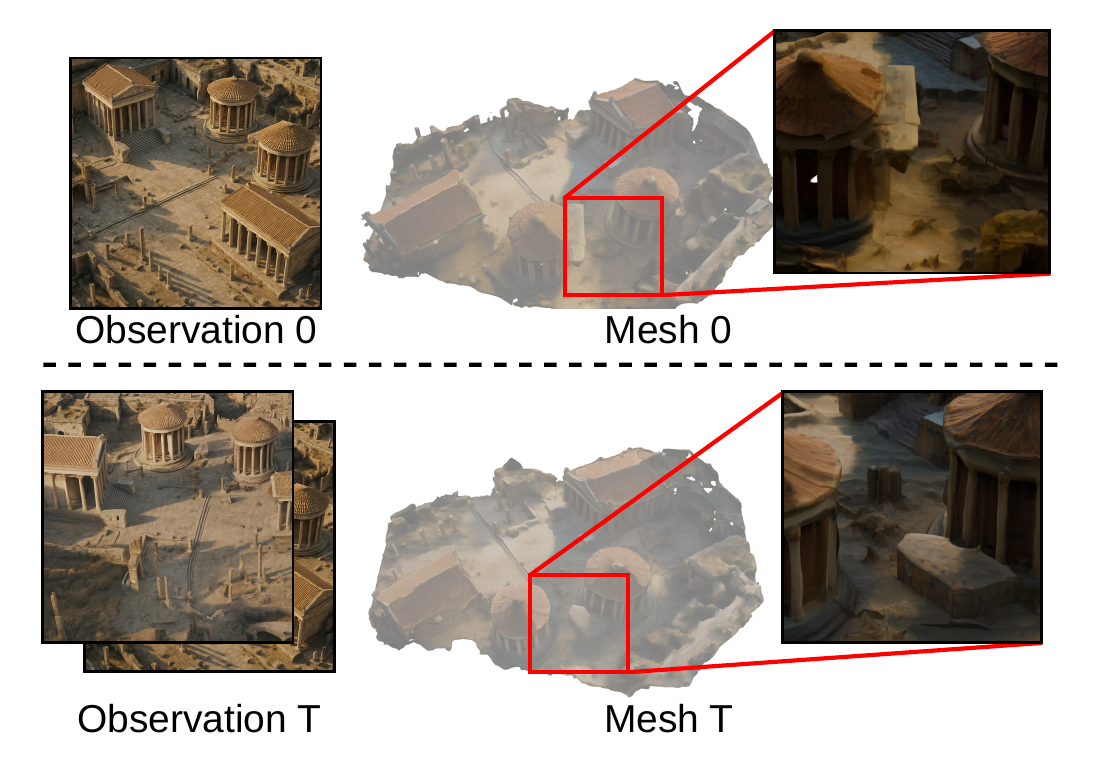}    
    \caption{\textbf{Ablation: Impact of iterative refinement.} Comparison between single-iteration (Mesh 0) and multi-iteration (Mesh T) reconstruction. The single-iteration mesh exhibits geometric errors in the highlighted region (distorted architectural structures), which are corrected after iterative refinement through accumulated multi-view observations and geometry-appearance co-evolution.}
    \label{fig:ablation_iterative}
    \vspace{-1.5em}
\end{figure}

\begin{figure}[t]
    \centering
    \includegraphics[width=\columnwidth]{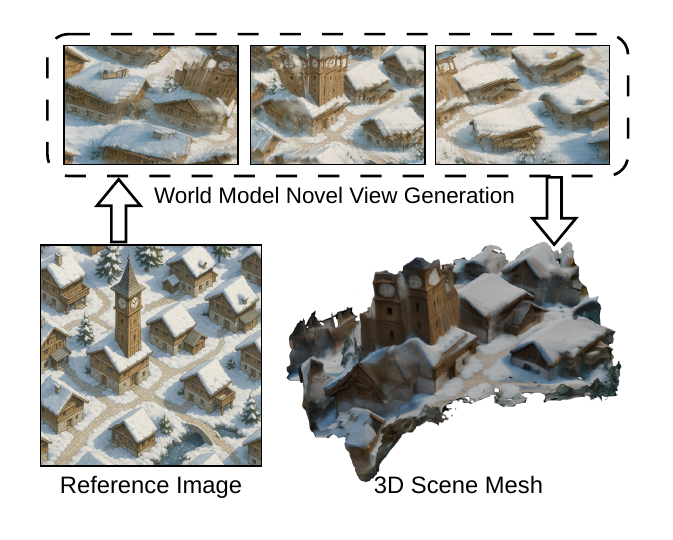}
     \vspace{-1.5em}
    \caption{\textbf{Ablation: Impact of depth-conditioned video guidance.} Comparison between pose-only world model (FlashWorld~\cite{li2025flashworld}) and our depth-conditioned video generation. The pose-only baseline generates visually plausible frames but produces geometrically inconsistent 3D meshes with severe distortions and broken surfaces (right). Our depth conditioning provides geometric scaffolding that ensures multi-view consistency and stable 3D reconstruction.}
    \label{fig:ablation_depth}
    \vspace{-1.5em}
\end{figure}

We conduct ablation studies to validate two core design choices: (1) iterative refinement versus single-iteration generation, and (2) depth-conditioned versus pose-only video generation. These ablations isolate the contributions of our self-evolving mechanism and geometric guidance. Quantitative results are in Table~\ref{table:ablation}, and qualitative comparisons in Figures~\ref{fig:ablation_iterative} and~\ref{fig:ablation_depth}.

\paragraph{Impact of Iterative Refinement.}
We compare single-iteration reconstruction (using only the input image) versus our full multi-iteration approach. Figure~\ref{fig:ablation_iterative} reveals critical differences: the single-iteration mesh (Mesh 0) exhibits noticeable geometric errors in the highlighted region, including distorted architectural structures and incomplete surfaces. This stems from limited viewpoint coverage, as a single image provides insufficient constraints to accurately reconstruct occluded regions. Through iterative refinement (Mesh T), our method progressively accumulates multi-view observations, providing increasingly rich geometric constraints. Each iteration's video generation reveals new viewpoints, which inform subsequent 3D generation stages, enabling the system to correct initial errors and complete previously unseen regions with high fidelity.

Table~\ref{table:ablation} quantifies this improvement with the full method achieving 54.0\%-56.5\% win rates in human evaluations and 58.2\%-60.5\% in GPT-4o assessments. While these margins may appear modest, they reflect the inherent challenge that even single-iteration results are reasonably plausible. The consistent preference for multi-iteration results across all three criteria (geometry, layout, texture) confirms that iterative geometry-appearance co-evolution is essential for achieving high-quality scene completion with accurate detail preservation.

\paragraph{Impact of Depth-Conditioned Video Guidance.}
We compare our depth-conditioned video generation against a pose-only world model baseline (FlashWorld~\cite{li2025flashworld} generating views from $-30^\circ$ to $+30^\circ$). Figure~\ref{fig:ablation_depth} reveals a striking disparity: while the pose-only model generates visually plausible individual frames, the reconstructed 3D mesh exhibits catastrophic geometric failures with severe distortions and broken surfaces. Without explicit geometric guidance, the model hallucinates appearance details that look plausible in 2D but violate 3D consistency constraints. Our depth-conditioned approach addresses this by rendering depth maps from the progressively refined mesh to guide video generation, providing geometric scaffolding that ensures multi-view consistency. Table~\ref{table:ablation} shows overwhelming win rates of 90.0\%-94.0\% in human evaluations and 97.9\%-99.1\% in GPT-4o assessments, confirming depth conditioning is absolutely critical for geometrically stable reconstruction under large viewpoint variations.  

\section{Conclusion}

We presented \modelname, a self-evolving framework for single-image 3D scene generation that progressively refines geometry and appearance through iterative cycles between 2D and 3D domains. By synergistically combining geometric knowledge from 3D generation models with visual priors from video generation models, our training-free approach addresses key limitations of prior single-pass methods. Through three coupled stages, spatial prior initialization, visual-guided mesh generation, and spatial-guided novel view synthesis, \modelname achieves complete full scene coverage with photorealistic textures. Comprehensive experiments demonstrate substantial improvements over state-of-the-art baselines, with human preference win rates reaching 78.5\%-90.5\% and ablations confirming the critical importance of both iterative refinement and depth-conditioned video guidance. We hope our work inspires further research into self-evolving approaches that bridge complementary strengths of different generative models for 3D content creation.

\clearpage
{
    \small
    \bibliographystyle{ieeenat_fullname}
    \bibliography{main}
}

\clearpage
\appendix
\begin{figure*}[!t]
     \centering
     \includegraphics[width=\textwidth]{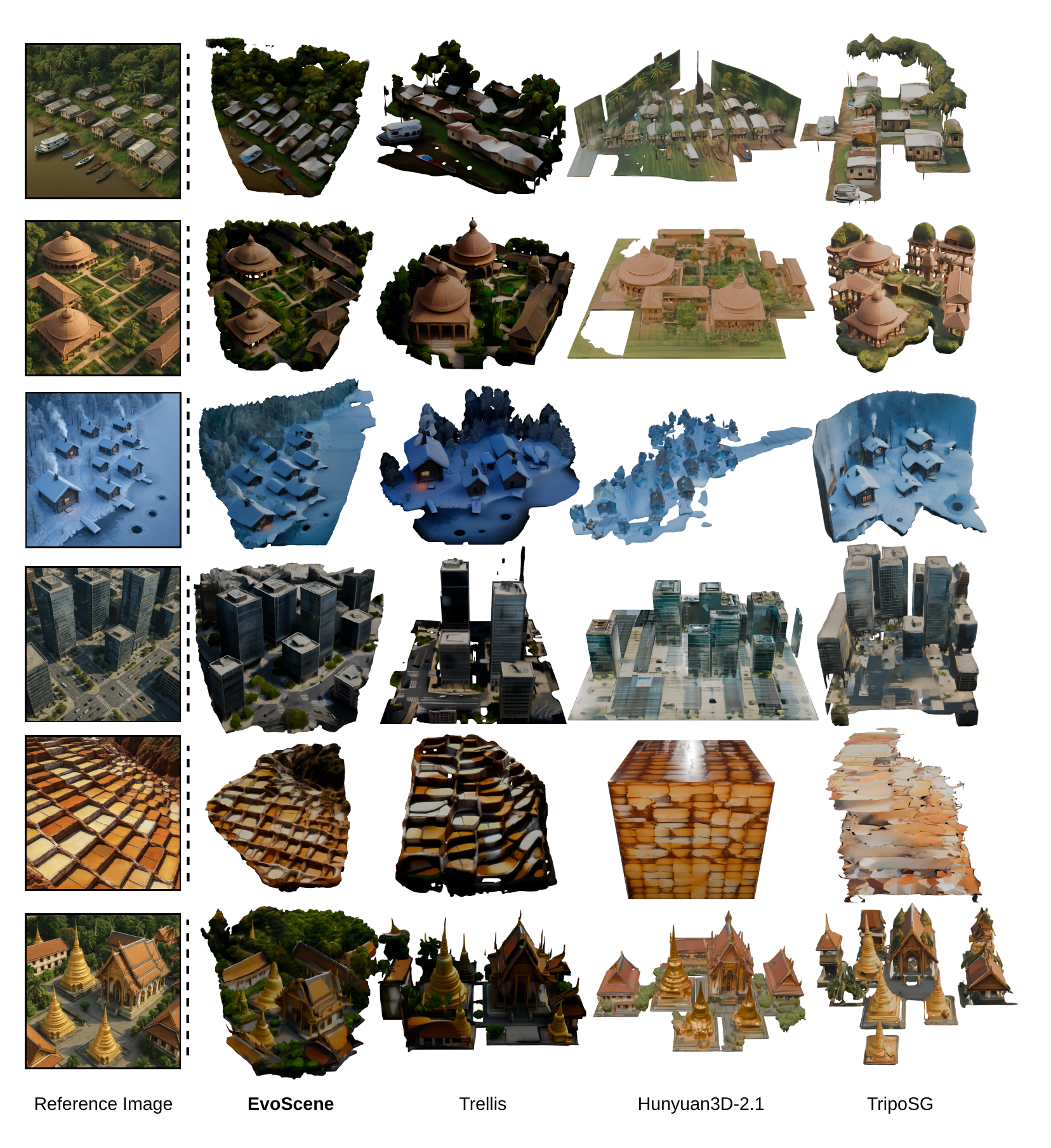}
     \caption{\textbf{Additional qualitative comparisons across diverse scene types.} From left to right: reference images, \modelname (ours), Trellis, Hunyuan3D-2.1, and TripoSG. \modelname produces complete geometry with preserved architectural details and photorealistic textures, while baselines exhibit fragmentation, flattened representations, or severe geometric distortions.}
     \label{fig:appendix_baseline_comparison}
\end{figure*}

\section{Method Details}
\label{sec:appendix_method}

\paragraph{Depth Estimation and Point Cloud Construction.}
We employ HunyuanWorld-Mirror~\cite{liu2025worldmirror} for metric depth estimation and camera parameter prediction from both single images and video frames. For the initial image $I_0$, the model predicts a metric depth map and camera intrinsics/extrinsics, which we back-project to obtain the initial point cloud $\mathcal{P}_0$. In subsequent iterations, we extract frames from generated videos and process each frame through HunyuanWorld-Mirror to obtain per-frame depth maps and camera parameters. To ensure geometric consistency, we apply multi-view geometric voting: for each candidate 3D point from frame $k$, we project it into all other frames and verify depth agreement within a tolerance of 0.1 meters. Points with support from at least 3 views are retained and assigned confidence scores based on the number of supporting views and local depth gradient magnitude (lower gradients indicate more reliable estimates). The filtered point clouds from all frames are merged with previous iterations' point clouds $\mathcal{P}_{t-1}$ through spatial binning at 5cm resolution, keeping points with highest confidence in each bin.

\paragraph{3D Scene Mesh Generation.}
We leverage Trellis~\cite{xiang2024structured}, a SLAT-based 3D diffusion model, to generate complete scene meshes. The accumulated point cloud $\mathcal{P}_t$ is voxelized at resolution $128^3$ to produce an occupancy grid $\mathcal{O}_t$ with three states: observed voxels (from point cloud), carved free-space (along viewing rays), and unknown occluded regions. For large scenes, we decompose the grid into overlapping $64^3$ patches with 48-voxel overlap and extract corresponding image crops from accumulated observations. We employ Trellis's two-stage flow-matching generator: the sparse structure generator first completes the binary occupancy. The structured latent generator then produces per-voxel latent features conditioned on the completed occupancy and image crops. During generation, we apply test-time optimization with multi-view rendering: every denoising steps, we decode the current SLAT into 3D Gaussians, render from all accumulated views, and compute gradients of the photometric loss $\mathcal{L}_{\text{mv}}$ (Eq.~\ref{eq:render}) with weights $\lambda_1=1.0$, $\lambda_2=1.0$, $\lambda_3=1.0$. These gradients guide 5 optimization steps (learning rate 1.0) before resuming diffusion. The final SLAT is decoded into both a mesh (via Marching Cubes) and 3D Gaussians for differentiable rendering.

\paragraph{Depth-Conditioned Video Generation.}
For novel view synthesis in Stage C, we render the completed mesh from $N=121$ viewpoints along an orbital trajectory spanning the target rotation angle. Camera trajectories alternate between iterations: iteration 1 uses azimuth range $[0^{\circ}, +45^{\circ}]$ and iteration 2 uses $[0^{\circ}, -45^{\circ}]$, with elevation and radius fixed to the camera pose of the original reference image. For each viewpoint, we render a disparity map (inverse depth) from the mesh, normalize it to $[0, 1]$, and resize to $1024\times1024$. These disparity maps serve as geometric conditioning for Wan2.2-Fun-5B-Control~\cite{wan2025}, a depth-conditioned video diffusion model. We construct the prompt as: "Camera orbiting around a static 3D scene, filling the blank space with detailed and realistic details, geometry consistent, high quality, 8k. The video shows [caption]", where the optional caption is generated by Qwen3-VL-2B~\cite{yang2025qwen3} from the input image. Video generation uses 50 DDIM sampling steps with classifier-free guidance scale 7.5, and the conditioning strength (controlnet scale) is set to 0.4 to balance geometric fidelity and visual quality. The input image $I_0$ is injected as the first frame to ensure appearance consistency.

\paragraph{Implementation and Computational Cost.}
All experiments are conducted on a system with one NVIDIA H100 (80GB) GPUs. Each pipeline iteration takes approximately 8 minutes: depth estimation and point cloud construction ($\sim$ 1 min), mesh generation with test-time optimization ($\sim$ 3 min), and video generation ($\sim$ 3 min). The complete three-iteration pipeline requires approximately 20 minutes per scene (No video generation for the last iteration). Peak GPU memory usage is approximately 68GB during mesh generation (for batch processing of $64^3$ patches) and 45GB during video generation. All models are used off-the-shelf without fine-tuning, demonstrating the training-free nature of our approach.

\begin{figure*}[!t]
     \centering
     \includegraphics[width=\textwidth]{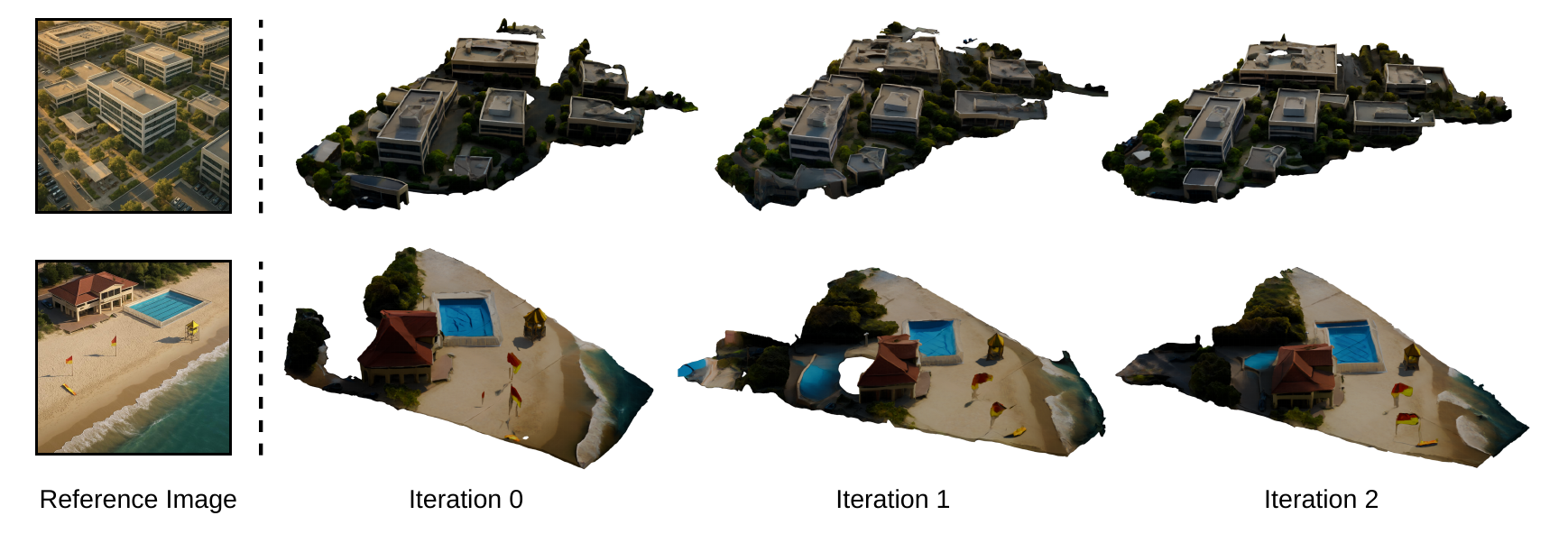}
     \caption{\textbf{Qualitative comparisons across different iterations.} The image shows the mesh differences between different iterations. With the iteration increasing, \modelname can progressively improve the geometry and appearance quality, especially for unseen regions within the reference images.}
     \label{fig:appendix_iteration_progression}
\end{figure*}
\section{More Experiment Results}
\label{sec:appendix_experiment}

\paragraph{Evaluation Metrics}

We employ both human evaluation and GPT-4o-based automatic evaluation to comprehensively assess 3D scene generation quality across three key dimensions: geometric quality, layout coherence, and texture consistency. Given a reference image and observations of two generated scenes (A and B), we ask annotators (or GPT-4o) to answer three questions: \emph{(i)} Which scene has geometry that is more detailed, precise, and closer to the reference image? \emph{(ii)} Which scene demonstrates a spatial layout and arrangement of objects that is more coherent and closely aligned with the layout in the reference image? \emph{(iii)} Which scene exhibits textures that are significantly more coherent and consistent with the reference image? For human evaluation, annotators select either A or B for each question, and we compute win rate as the percentage of times our method is preferred. For GPT-4o-based evaluation, we prompt the model to directly return the answer and extract top-5 token log probabilities, obtaining $P(\text{A})$ and $P(\text{B})$ (set to 0 if not in top-5). When our method is option A, we treat $P(\text{A})$ as a soft vote; when our method is option B, we use $P(\text{B})$. The weighted win rate is computed as $P(\text{win})$ if our method wins, and $1-P(\text{lose})$ if it loses, then averaged across all comparisons to capture both preference frequency and confidence strength.

\paragraph{Extended Baseline Comparisons}
Figure~\ref{fig:appendix_baseline_comparison} presents additional qualitative comparisons between \modelname and three strong baselines (Trellis~\cite{xiang2024structured}, Hunyuan3D-2.1~\cite{hunyuan3d22025tencent}, and TripoSG~\cite{li2025triposg}) across diverse scene categories including urban environments, natural landscapes, architectural landmarks, and residential areas. As discussed in the main paper (Section 4.2), \modelname consistently produces complete geometry with preserved architectural details and photorealistic textures, while baselines exhibit fragmentation (TripoSG), flattened representations (Hunyuan3D-2.1), or incomplete geometry with missing regions (Trellis). These additional examples further validate our method's superior performance across varying scene complexity and demonstrate robust generalization to diverse visual content.

\paragraph{Iterative Refinement Visualization}
Figure~\ref{fig:appendix_iteration_progression} illustrates the progressive improvement of geometry and appearance through our three-iteration self-evolving process (iterations 0, 1, 2). The initial iteration 0 reconstruction from only the input image produces reasonable but incomplete meshes with geometric errors in occluded regions and limited texture detail. Iteration 1 adds synthesized views from azimuth range $[0^{\circ}, +45^{\circ}]$, revealing previously occluded regions and improving background structure accuracy. Iteration 2 synthesizes complementary views from $[0^{\circ}, -45^{\circ}]$, achieving near-complete coverage with substantially refined geometric accuracy and texture consistency across the entire scene. The progression clearly demonstrates how our iterative geometry-appearance co-evolution progressively corrects initial errors, expands spatial coverage, and enhances both structural accuracy and visual realism, with the largest improvements occurring between iterations 0 and 1 when multi-view observations dramatically expand scene knowledge.

\end{document}